\documentclass[runningheads]{llncs}

\usepackage[T1]{fontenc}
\usepackage{cite}
\usepackage{amsmath,amssymb,amsfonts}
\usepackage{url}
\usepackage{algorithmic}
\usepackage{graphicx}
\usepackage{textcomp}
\usepackage{xcolor}
\usepackage{glossaries}
\usepackage{listings}
\usepackage{url}
\usepackage{enumerate}
\usepackage[normalem]{ulem}
\usepackage{bbding}

\newcommand{\atomicadd}{\texttt{atomicAdd}}

\newcommand{\kRO}{$K_{\textsc{RO}}\ $}

\newcommand{\kRORDGEMM}{$K_{\textsc{RO-RDGEM}}$}
\DeclareMathOperator*{\argmax}{arg\,max}

\begin{document}
\newacronym{HPC}{HPC}{high-performance computing}
\newacronym{DL}{DL}{Deep Learning}
\newacronym{ML}{ML}{Machine Learning}
\newacronym{AI}{AI}{artificial intelligence}
\newacronym{DOE}{DOE}{US Department of Energy}
\newacronym{DNN}{DNN}{deep neural network}
\newacronym{DNNs}{DNNs}{deep neural networks}
\newacronym{DFT}{DFT}{density functional theory}
\newacronym{FPNA}{FPNA}{floating-point non-associativity}
\newacronym{APFPR}{APFPR}{Asynchronous Parallel Floating Point Reductions}
\newacronym{GPU}{GPU}{graphics processing unit}
\newacronym{FP}{FP}{floating-point numbers}
\newacronym{FP64}{FP64}{double precision floating-point numbers}
\newacronym{FP32}{FP32}{single precision floating-point numbers}
\newacronym{FP16}{FP16}{half precision floating-point numbers}
\newacronym{FP8}{FP8}{half half precision floating-point numbers}
\newacronym{FP4}{FP4}{half half half precision floating-point numbers}
\newacronym{PDF}{PDF}{probability density function}
\newacronym{SPSA}{SPA}{simple-pass-with \texttt{atomicAdd}}
\newacronym{AO}{AO}{\texttt{atomicAdd}-only}
\newacronym{CU}{CU}{sum function from the CUB/HIPCUB library}
\newacronym{SPTR}{SPTR}{single-pass-with-tree-reduction}
\newacronym{FIFO}{FIFO}{first in first out queue}
\newacronym{TPRC}{TPRC}{two-passes-with-final-reduction-on-CPU}
\newacronym{SPRG}{SPRG}{single-pass-with-final-recursive-sum-on-GPU}
\newacronym{KL}{KL}{Kullback–Leibler divergence criterion}
\newacronym{GNN}{GNN}{graph neural network}
\newacronym{GNNs}{GNNs}{Graph neural networks}
\newacronym{DGEMM}{DGEMM}{double-precision matrix-matrix multiplication}
\newacronym{RDGEMM}{RDGEMM}{with a DGEMM running in a different CUDA stream}
\newacronym{RO}{RO}{reduction only}
\newacronym{FGSM}{FGSM}{fast gradient signed attack}
\newacronym{BIEO}{BIEO}{block index {\it vs} execution order}
\newacronym{PGD}{PGD}{projected gradient descent}
\newacronym{LLM}{LLM}{Large Language Models}
\newacronym{NN}{NN}{Neural Network}
\newacronym{EWA}{EWA}{external workload attack}
\newacronym{MiG}{MiG}{multi-instance GPU}
\newacronym{LP}{LP}{Learnable Permutation}

%\title{The effects of floating-point associativity and asynchronous parallel execution on the robustness of machine learning classifications.}

\title{Robustness of Deep Learning Classification 
%to floating-point non-associativity 
to Adversarial Input on GPUs: Asynchronous Parallel Accumulation is a Source of Vulnerability}\vspace{-1mm}
\titlerunning{Asynchronous Parallel Accumulation as Vulnerability}

\author{Sanjif Shanmugavelu\inst{1}\and %\orcidID{0009-0003-8245-4013} 
Mathieu Taillefumier\inst{2} \and %\orcid{0000-0002-3983-5625}
Christopher Culver\inst{1} \and
Vijay Ganesh\inst{3}\and%\orcid{0000-0002-6029-2047}  
Oscar Hernandez\inst{4} \and
Ada Sedova\inst{4} %0000-0002-8233-3057
} 
\institute{Maxeler Technologies, a Groq Company. 3 Hammersmith Grove, London, UK \and
    ETH Zurich / CSCS, OAT V floor, Andreasstrasse 5, 8092 Zurich, CH\and
    Georgia Institute of Technology, Atlanta, USA\and
    Oak Ridge National Laboratory. Oak Ridge, TN, USA\\
\email{sshanmugavelu@groq.com\Envelope, 
tmathieu@ethz.ch, sedovaaa@ornl.gov}}
\authorrunning{Shanmugavelu et al.}
\maketitle 

%\vspace{-1em}

\begin{abstract}

% Oscar comment

% The attack can also be encoded on hw. Like producing a hw that triggers something on a model.

The ability of machine learning (ML) classification models to resist small, targeted input perturbations---known as adversarial attacks---is a key measure of their safety and reliability. 
%We show here that floating-point non-associativity (FPNA), coupled with asynchronous parallel programming on GPUs, is sufficient to result in model misclassification \textit{without any perturbation to the input}. Misclassification is particularly significant for inputs close to the decision boundary, and standard adversarial robustness results may be overestimated up to 4.6\% when not considering machine-level details. 
We show that floating-point non associativity (FPNA) coupled with asynchronous parallel programming on GPUs is sufficient to result in misclassification, \textit{without any perturbation to the input}. Additionally, we show that this misclassification is particularly significant for inputs close to the decision boundary and that standard adversarial robustness results may be overestimated up to 4.6 when not considering machine-level details.
We first study a linear classifier, before focusing on standard Graph Neural Network (GNN) architectures and datasets used in robustness assessments. 
%We develop a novel black-box attack using Bayesian optimization to discover external workloads that can bias the output of reductions on GPUs by changing scheduler ordering and \textit{reliably} lead to misclassification via this vulnerability. 
We develop a novel black-box attack using Bayesian optimization to discover external workloads that can change the instruction scheduling which bias the output of reductions on GPUs and \textit{reliably} lead to misclassification. Motivated by these results, we present a new learnable permutation (LP) gradient-based approach to learning floating-point operation orderings that lead to misclassifications. The LP approach provides a \textit{worst-case} estimate in a computationally efficient manner, avoiding the need to run identical experiments tens of thousands of times over a potentially large set of possible GPU states or architectures. Finally, using instrumentation-based testing, we investigate parallel reduction ordering across different GPU architectures under external background workloads, when utilizing multi-GPU virtualization, and when applying power capping. Our results demonstrate that parallel reduction ordering varies significantly across architectures under the first two conditions, substantially increasing the search space required to fully test the effects of this parallel scheduler-based vulnerability. These results and the methods developed here can help to include machine-level considerations into adversarial robustness assessments, which can make a difference in safety and mission critical applications.%\vspace{-1em}

% \keywords{Deep learning \and machine learning \and sensitivity \and reproducibility \and floating-point arithmetic \and parallel programming \and high-performance computing}
\end{abstract}%\vspace{-1em}

\section{Introduction}%\vspace{-1mm}

\gls{DL} models are increasingly used in safety-critical applications such as autonomous vehicles, medical diagnostics, and laboratory automation, where reliability and robustness are crucial~\cite{perez-cerrolaza2023survey, rabbani2022laboratory, szymanski2023synthesis}.
%,brix2023fourth
Their growth has been fueled by hardware accelerators such as graphics processing units (GPUs) that enable high-throughput training and deployment~\cite{youvan2023parallelprecision}. %, pandey2022gpudrugdiscovery, youvan2023parallelprecision}. %, 1575717, jouppi2017indatacenterperformanceanalysistensor. 
 As \gls{ML} models gain traction in safety-critical applications, ensuring their robustness is essential. 
 
 A key metric is robustness to adversarial attacks—crafted perturbations that induce misclassification, often generated via gradient-based methods ~\cite{szegedy2014intriguingpropertiesneuralnetworks, goodfellow2014explaining, madry2018towards}. These methods maximize prediction error by modifying inputs while keeping model parameters fixed. 
While robustness efforts focus on hyperparameter tuning, model architecture, and adversarial training, verification tools often overlook system and machine-level fluctuations and \gls{FPNA} in parallel computing~\cite{summers2021nondeterminism}. However, hardware attacks such as side-channel exploits, hardware Trojans, and fault injection attacks represent an expanding area of concern, further threatening the security and reliability of \gls{DL} \cite{Gaine2023,Lee2022}. %,Ashraful2023
% Distributed compute and optimizations such as atomic operations and minifloat formats have additionally enabled the support of larger datasets and model parameters, further improving performance. 
The demand for compute has expanded the market for GPU-based cloud services, with providers like AWS, Azure, and GCP offering on-demand resources. Additionally, new accelerators like the Groq LPU and Cerebras WSE address GPU bottlenecks, an increase the diversity of hardware used. Popular \gls{ML} models, including recommendation systems and \gls{LLM}, are often delivered via APIs in a Models as a Service (MaaS) framework, with users having little knowledge of low-level details such as parallel reduction schemes. Factors such as virtualization, background workloads, power capping, and floating-point precision are often not disclosed, making it challenging to understand their impact on model performance and robustness. 
%\vspace{-1mm}

\textbf{Limitations of State-of-The-Art:} 
 We introduce the term \gls{APFPR} to define the problem of run-to-run variability due to the combination of \gls{FPNA} and asynchronous parallel programming. 
Previous work~\cite{shanmugavelu2024impactsfloatingpointnonassociativityreproducibility} analyzed this effect in PyTorch functions and \gls{DL} models, notably GNNs. However, its impact on misclassification---crucial for accuracy and robustness---remains an open question. Here, we investigate how \gls{APFPR}-induced variability leads to model misclassification. To our knowledge, existing methods evaluating robustness do not consider machine-level factors; in particular, \textit{they do not consider hardware-level fluctuations and how these could be exploited as attacks}. While the issue also applies identically to CPUs, here we focus on GPUs due to their widespread use in \gls{ML}. We highlight our main contributions as follows. All codes and artifacts are made available at \url{https://www.github.com/minnervva/fpna-robustness}. 

\textbf{(1) Machine-Induced Misclassification of Fixed Inputs:} 
Misclassifications do not always require input perturbations; they may arise from \gls{APFPR} on GPUs. Asynchronous programming is often used by default in \gls{DL} programs on GPUs via atomic operations with unspecified execution orders \cite{d9m}. We show that adversarial robustness is vulnerable to \gls{APFPR}. Misclassifications may only occur after thousands of identical runs, therefore, exhaustive searches to rigorously characterize robustness due to \gls{APFPR} are required and are impractical, highlighting the need for analytical or heuristic approaches. These results apply across frameworks such as PyTorch, TensorFlow, and JAX \cite{d9m}. 
%\vspace{-0.5mm}

\textbf{(2) External Workload Attack:} 
We introduce a black-box \gls{EWA} that uses Bayesian optimization to identify workload properties leading to misclassification via the reordering of \gls{APFPR} operations, requiring \emph{only} knowledge of possible output classes. We focus on external workloads involving matrix multiplication, optimizing the matrix size to induce misclassification of a fixed input.
%\vspace{-0.5mm}

\textbf{(3) Learnable Permutations to Estimate Worst-Case Robustness:} We propose a heuristic gradient-based method to identify permutations that induce misclassification, providing a worst-case robustness estimate. This approach eliminates the need for multiple identical iterations on a fixed input and generalizes across all possible GPU states. If GPU scheduling details were available, the method could also be used as an attack.
%\vspace{-0.5mm}

\textbf{(4) Benchmarks:}
We investigate robustness on the standard robustness-assessment GNN datasets and models, highlighting their vulnerability due to a non-deterministic base class \cite{Torch-Documentation}. We show that a \gls{EWA} can reliably induce misclassification and that the learnable permutation approach provides a tight upper bound on robustness. 
%While our analysis focuses on GNNs, the results and approach generalize to other architectures.
%\vspace{-0.5mm}

\textbf{(5) Impact of GPU State on Reduction Ordering:} 
Using the asynchronous parallel sum as a test, we track the execution order of atomic operations relative to the block index using source-code instrumentation. We reveal execution order variations across GPU architectures, testing the state under virtualization, background workloads, and power capping. While GPU virtualization and external workloads significantly affect instruction ordering, we find that power capping has little impact.\vspace{-4mm}

% do we need this sentence.
%\vspace{-4mm}

%%%%%%%%%%%%%%%%%%%%%%%%%%%%%%%%%%%%%%%%%%%%%%%%%%%%%%%%%%%%
\section{Impacts of \gls{APFPR} Non-Determinism on Classification}%\vspace{-3mm}
%%%%%%%%%%%%%%%%%%%%%%%%%%%%%%%%%%%%%%%%%%%%%%%%%%%%%%%%%%%%

\begin{figure}[htbp]
\centering
\includegraphics[width=0.495\textwidth]{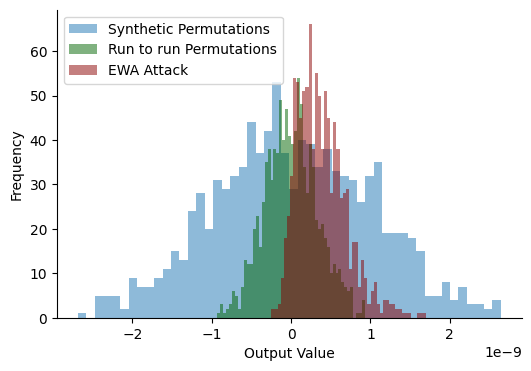}
\includegraphics[width=0.495\textwidth]{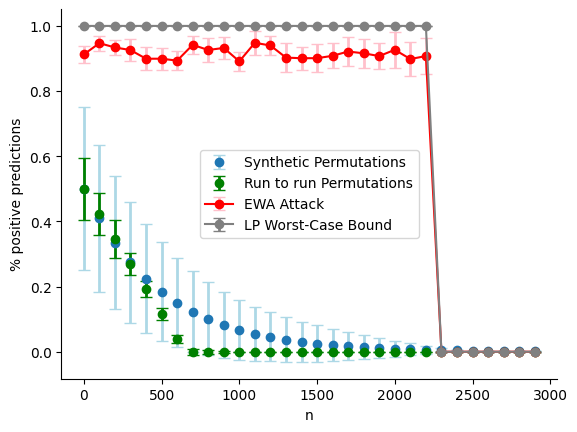}
\caption{  
  \textbf{Left panel:} Probability density of the output \(\hat{\mathbf{n}} \cdot \mathbf{x}\) which has a theoretical value of 0. Both vectors \(\hat{\mathbf{n}}\) and \(\mathbf{x}\) have dimensionality \(d = 1,000\). 
  \textbf{Right panel:} Analysis of points on the decision boundary \(\hat{\mathbf{n}} \cdot \mathbf{x} = b\) with iterative perturbations of the form \(n \cdot \epsilon + \hat{\mathbf{n}}\), where \(\epsilon = 1 \times 10^{-12}\) and \(0 \leq n \leq 3000\). Experiments performed on the H100 with FP64 precision. We consider synthetic and real permutations (Section \ref{sec:linear-classifier-synthetic}) in addition to \gls{EWA} attacks and LP worst-case bounds (Section \ref{sec:ewa}).
}

\label{fig:linear_decision_boundary_distribution}
\label{fig:linear_decision_boundary_adversarial_attack}
\end{figure}

To better understand the impact of \gls{APFPR} on classifier robustness, we first construct synthetic examples, manually introducing permutations in the order of floating-point reductions, before considering real run-to-run variability. 
%Our goal is to reveal and quantify the extent to which non-deterministic parallel programming can lead to misclassification, despite using identical models and inputs.
In this section, we develop a simple classifier with a linear decision boundary and use it to investigate the misclassification of points close to the boundary. We demonstrate the inherent difficulty of exploring the combinatorial space of permutations: brute-force testing may not fully explore those that result in misclassification. The \gls{EWA} we then introduce can reliably induce misclassifications on a fixed input by running an external workload, thereby altering the ordering of asynchronous operations on the GPU.
%\sout{by the scheduler}
Finally, we introduce a \gls{LP} scheme that models reduction orderings in asynchronous programming with a differentiable representation. Using a heuristic gradient-based optimizer, we efficiently identify permutations that maximize predictive error, providing a systematic approach to identify inputs susceptible to \gls{EWA}. Fig.~\ref{fig:linear_decision_boundary_adversarial_attack} %\textcolor{red}
%{\sout{displays the results of all experiments performed in the analysis portions of the following subsections, and is referred to accordingly} 
summarizes all results discussed in the paper. %First, we introduce terminology and definitions used throughout the paper.

%\vspace{-1em}

\subsection{Theoretical Description of Permutation-Based Misclassification}\label{sec:defs}%\vspace{-1mm}

%We show that the complexity of the function describing the decision boundary makes it sensitive to \gls{APFPR} variability.

Let $f : \mathbb{R}^{d} \longrightarrow \mathbb{R}^{L}$,  $d, L \in \mathbb{N}$, be an arbitrary multiclass classifier where $L$ is the number of classes. Given a datapoint $\mathbf{x} \in \mathbb{R}^{d}$, the class $1 \dots L$, which the classifier $f$ predicts for $\mathbf{x}$, is given by $\hat{k}(\mathbf{x}) = \operatorname{\argmax}_{i} {f}_{i}(\mathbf{x})$, where ${f}_{i}(\mathbf{x})$ is the $i$-th component of an array of probabilities called logits.
We define a function $F$ at point $\mathbf{x} \in \mathbf{R}^{d}$ by 
\begin{equation}
   F(\mathbf{x}) = \mathrm{max}_{i}{f}_{i}(\mathbf{x}) - \mathrm{max}_{i \neq \hat{k}(\mathbf{x})} {f}_{i}(\mathbf{x})
\end{equation}
$F$ describes the difference between the likelihood of classification for the most probable and the second most probable class. For a given $\mathbf{x} \in \mathbf{R}^{d}$, the higher the value of $F(\mathbf{x})$, the more confident we are in the prediction given by the classifier. The decision boundary $B$ of a classifier $f$ can then be defined as the set of points $\mathbf{x}$ that are equally likely to classify into at least two distinct classes:
\begin{equation}
    B = \{\mathbf{x} \in \mathbb{R}^{d} :  F(\mathbf{x}) = 0\}
\end{equation}
$B$ splits the domain $\mathbb{R}^{d}$ into subspaces of similar classification.
Given $\mathbf{x} \in \mathbb{R}^{d}$ and a perturbation $\delta({\mathbf{x}}) \in \mathbb{R}^{d}$ such that $\mathbf{x} + \delta({\mathbf{x}}) \in B$, we have that $\mathbf{x} + \delta({\mathbf{x}})$ is on the boundary of misclassification. Hence, when considering misclassification, we study the properties of the decision boundary $B$. In the following, we study the properties of $B$ under a perturbation $\delta x\in\mathbb{R}^{d}$ around $\mathbf{x} \in \mathbb{R}^{d}$, called an adversarial attack under the constraint $\mathbf{x} + \delta({\mathbf{x}}) \in B$.

Adversarial attacks are small, often imperceptible changes made to input data $\mathbf{x}_{adv} = \mathbf{x} + \delta(\mathbf{x})$, that cause the model to misclassify. 
%These perturbations can be viewed as modifications to the input that affect the model's decision boundary $B$.
Among the most notable adversarial attacks are the gradient-based \gls{FGSM}~\cite{goodfellow2014explaining} and \gls{PGD}~\cite{madry2018towards} attacks. \gls{FGSM} generates adversarial examples by adding perturbations in the direction of the gradient of the loss function, while \gls{PGD}  is based on an iterative application of \gls{FGSM} ~\cite{madry2018towards}. We also consider a variant of margin-based attacks~\cite{Margin}, which we call a \emph{targeted attack}. This targeted attack finds examples closer to the decision boundary $B$ by minimizing the function $F$.
%decreasing the difference in the likelihood of classification of the most and second most likely class. 
Random attacks that add random noise to the inputs provide a baseline for evaluating the model's robustness. In general, these attacks are evaluated with an attack scale factor $\epsilon$, such that $\mathbf{x}_{adv} = \mathbf{x} + \epsilon \cdot \boldsymbol{\delta}(\mathbf{x})$ where $\epsilon \geq 0$.%\vspace{-2mm}

\subsection{Synthetic and Real Permutations}%\vspace{-1mm}

\subsubsection{Simple Linear Classifier: Examination of the Decision Boundary:} \label{sec:linear-classifier-synthetic}

We define a linear classifier by its hyperplane decision boundary:  
\begin{equation}
    f : \mathbf{x} \in \mathbb{R}^{d} \longrightarrow \mathbf{1}\{ \hat{\mathbf{n}} \cdot \mathbf{x} \geq b \}
\end{equation} \label{eqn:classifier}
where $\hat{\mathbf{n}} \in \mathbb{R}^{d}$ is the normal vector to the hyperplane, $b \in \mathbb{R}$ is the bias, and $\mathbf{1}$ is the indicator function. This classifier assigns inputs to one of two classes based on whether they lie above or below the hyperplane defined by $\hat{\mathbf{n}} \cdot \mathbf{x} = b$. Using the notation from Section~\ref{sec:defs}, we express the decision boundary $B$ as:  
\begin{equation}
    B = \{\mathbf{x} : \hat{\mathbf{n}} \cdot \mathbf{x} - b = 0\}
\label{eqn:linear_example_solution}
\end{equation}
% The boundary $B$ is mathematically invariant to any permutation $\pi$ of the elements $x_i n_i$ in the dot product ${\bf x}\cdot {\bf n}=\sum_{i} x_i n_i$. 
Mathematically, the boundary $B$ inherits its invariance against a permutation $\pi$ of the elements $x_i n_i$ from the properties of the dot product. However, in practice, run-to-run variability in the decision boundary may arise due to \gls{APFPR}. To simulate the effects of \gls{APFPR}, we iterate over all possible permutations of the input and normal vector, then compute Eq.~\ref{eqn:classifier} for points $\mathbf{x}$ on the decision boundary $B$ with $b=0$. The points $\mathbf{x}$ are sampled from a normal distribution centered at the origin. 
%The resulting distribution of values for $d=1000$ and $\hat{\mathbf{n}} = \frac{1}{\sqrt{d(d-1)}} \cdot (d-1, -1, \dots, -1)$ is shown in Fig.~\ref{fig:linear_decision_boundary_distribution} (\emph{synthetic permutations} curves). We construct the normal vector as such to reduce the space of permutations from $d!$ to $d$.
Results for $d=1000$ and $\hat{\mathbf{n}} = \frac{1}{\sqrt{d(d-1)}} \cdot (d-1, -1, \dots, -1)$ (reducing the search space from $d!$ to $d$) are shown in Fig.~\ref{fig:linear_decision_boundary_distribution} (\emph{Synthetic Permutations}).

The observed distribution exhibits a spread around zero, with a minimum and maximum variation of approximately $\pm 3 \times 10^{-9}$. For the real-life case, we perform $N=1000$ identical runs for a fixed input. This distribution has a minimum and maximum variation of approximately $\pm 0.9 \times 10^{-9}$. Since the input is fixed and the only source of variation is the accumulation order of floating-point operations, we conclude that \gls{APFPR} can shift the decision boundary $B$ of  the classifier. Furthermore, we show that the set of permutations explored in real-life identical-runs (\emph{run-to-run permutations} curves in Fig. \ref{fig:linear_decision_boundary_adversarial_attack}) may not cover the set of all possible permutations. Misclassifications may occur as infrequently as once a thousand identical runs.

To study the effect of input perturbations on classifier's robustness, we consider points on the decision boundary $\hat{\mathbf{n}} \cdot \mathbf{x} = b$ and introduce deviations $n \cdot \epsilon + \hat{\mathbf{n}}$, where $\epsilon = 1 \times 10^{-12}$ and $0 \leq n \leq 3000$. As shown in the right panel of Fig.~\ref{fig:linear_decision_boundary_adversarial_attack}, when looping through all possible permutations, classification flips decreases with increasing $n$, and zero out at $n=2400$. 
We observe similar results for the percentage of positive predictions with a different $n=700$ in the real-life case ($N=1000$ identical runs). These results show that \gls{APFPR} cannot affect classification for inputs far enough from the boundary, and also that, repeated, identical real-life runs may not be sufficient to describe robustness, assuming all possible permutations can be explored at runtime. To explore these ideas further, Section \ref{subsec:LP} provides a heuristic to identify such ``safe'' points, while Section \ref{sec:ewa} describes a method to systematically find permutations that consistently induce misclassification through manipulation of system conditions. These findings can be generalized to other floating point formats.

% \textcolor{red}{We conclude that \gls{APFPR} cannot affect classification results for inputs far enough from the boundary. We provide a heuristic to identify such "safe" points in Section \ref{subsec:LP}. For the real-life case, we perform $N=1000$ identical runs with similar results, but the percentage of positive predictions zero out much sooner at $n=700$. The synthetic experiments Results show that the repeated identical real-life runs may not be sufficient to describe robustness, assuming all possible permutations may be explored at runtime. The above experiments exhibit similar behavior for FP16, FP32 and FP64 formats, with larger bounds at lower precisions. In Section \ref{sec:ewa}, We demonstrate a method to systematically manipulate system conditions, allowing exploration of permutations that consistently induce misclassification.}
%we show how to attack a system to explore permutations that result in misclassification.\vspace{-1mm}

\subsection{External Workload Attacks and Learnable Permutations} 

% We introduce a novel attack method to induce misclassifications on fixed inputs. Additionally, we present a defence technique with learnable permutations to identify inputs that are susceptible to this attack.

\subsubsection{External Workload Attacks} \label{sec:ewa}

We examine the linear decision boundary in Eq. \ref{eqn:linear_example_solution} under asynchronous computation on the H100, V100, and Mi250 GPUs. We introduce \gls{EWA}, which exploits the impact of background workloads on classification. As studied in detail in Section~\ref{sec:gpu-state}, additional workloads running on the same GPU as inference tasks can affect the ordering of \gls{APFPR}. Without loss of generality, we use square matrix multiplications as the background workload
%\textcolor{red}{\sout{, running in a separate process,}} 
and determine the optimal matrix size $k$ that reliably skews classifier outputs.

We use Bayesian optimization with the objective $O(k) = \mathbb{E}\left[\mathbf{1}\left(f(x, k), o\right)\right]$, where $o \in \{0, 1\}$ is the target output and $\mathbf{1}$ is the indicator function. We run the optimization for 100 iterations, with 1000 experiments per iteration, to find the optimal matrix size $k \in \{1000, 10000\}$ to flip the classifications into positive classes. Then, we perform 1000 repeated inferences to test the success of the attack. As shown in the right panel of Fig.~\ref{fig:linear_decision_boundary_adversarial_attack}, all inputs can be reliably skewed toward the desired classification at least 82.7\% of the time. The left panel of Fig.~\ref{fig:linear_decision_boundary_adversarial_attack} shows the positive skewed distribution. The \gls{EWA} attack is ineffective at $n=2400$, because no possible configuration of floating-point operations results in misclassification (see Section \ref{sec:linear-classifier-synthetic});
%The \gls{EWA}-induced errors are present from $n=0$ to $n=2400$, resulting in classification flips for the chosen FP64 format, after which they disappear, 
this illustrates the effectiveness of the Bayesian approach in finding a workload that can exploit any \gls{APFPR}-based vulnerability. We observe similar behavior with FP16 and FP32 formats.

We find that the \gls{EWA} optimization convergence behaves similarly across GPU families, although the optimal matrix size depends on the GPU family (our GitHub repository contains results for the other GPU architectures and datatypes). %\textcolor{red}{\sout{The trend in inputs and optimal attack matrix size}}
The relationship between the input and the optimal matrix size is erratic and we leave an in-depth investigation to future work as it would require developing tools to probe the GPU scheduler, which to our knowledge do not exist \cite{otterness2021exploring, Olmedo2020-nvidia-scheduler}. We note that \gls{EWA} may be inadvertently triggered in cloud systems where GPUs are virtualized and shared. Section \ref{sec:gpu-state} further explores this idea  by measuring the difference in the scheduling of atomic instructions in reductions using black-box testing, both with and without external workloads and analyzing other GPU state factors, including partitioning and power capping.
%\vspace{-1em}
\subsubsection{Learnable Permutation to Find Possible Adversarial Perturbations:}\label{subsec:LP}

We developed a gradient-based optimization technique to find a permutation of floating-point operations that causes misclassification. Following Section~\ref{sec:defs}, let $f$ be a classifier mapping an input tensor $\mathbf{x}$ to logits a probability vector of length $L$, the number of classes; in some cases $f$ may be composed of multiple functions $f_i$ with the same properties. We take the argmax of the logits to obtain the final classification. We require $f$ to include floating-point accumulations. Due to \gls{APFPR}, the output of $f$ depends on a set of permutation matrices $P_i$ describing the order of reductions, written as $f(P_i, \mathbf{x})$, where $i$ is the index for each permutation matrix associated to the functions $f_i$ composing the function $f$. 
%\textcolor{red}{\sout{In cases where $f$ is composed of multiple functions with the same properties, we parameterize using a set of permutation matrices $\{P_i\}$}}. 
The classifier is trained by minimizing a loss function $\mathcal{L}(f, y)$, where $y$ is the ground truth label. To find $\{P_i\}$ that cause $f$ to misclassify $\mathbf{x}$, we maximize the prediction error with respect to the permutation perturbation:

\begin{equation}
\begin{aligned}
& \text{maximize} && \mathcal{L}(f(\{P_i\}, \mathbf{x}), y) \\
& \text{subject to} && P_i^T P_i = I, \quad \, i = 1, 2, \ldots, L
\end{aligned} \label{en:LP_optimization}
\end{equation}
We use the Gumbel-Softmax technique \cite{mena2018learninglatentpermutationsgumbelsinkhorn, Jang2017-gumbel-softmax} %maddison2017concretedistributioncontinuousrelaxation
to create a differentiable approximation of the permutation matrices. By adding Gumbel noise and applying softmax to the matrix, we can use gradient descent to optimize the set $\{P_i\}$ that maximizes the loss function, with the other parameters of $f$ fixed. Next, valid permutation matrices are obtained by solving the linear assignment problem via the Hungarian algorithm \cite{Kuhn1955-hungarian-method}. This method, inspired by adversarial attacks (Section~\ref{sec:defs}), maximizes error with respect to floating-point operation ordering instead of the input. While the approach does not guarantee misclassification, it provides a more efficient way to find adversarial permutations compared to brute-force search.

We now present practical steps to use the LP method: (1) Identify non-deterministic functions by referencing documentation or using a linter like the \emph{torchdet} tool \cite{TorchDet}, (2) For each non-deterministic function, introduce a permutation matrix $P$ to simulate runtime variations in reductions. For example, consider the linear classifier in Sec.~\ref{sec:linear-classifier-synthetic}. In that case, we compute $P \hat{\mathbf{n}} \cdot P \mathbf{x}$ instead of $\hat{\mathbf{n}} \cdot \mathbf{x}$. For a fully connected linear layer with a weight matrix $w$ of size $N \times M$ and a bias vector $\mathbf{b}$ of size $N$, where the intermediate output \( \mathbf{y} \) is given by
$\mathbf{y} = w^{T} \mathbf{x} + \mathbf{b}$, we apply the permutation matrix $P$ on element-wise products $S_{i}$ before reduction where \( S_i = \{w_{i0}x_0, \dots, w_{iM}x_M\} \), computing  $y_i = \sum_{j=1}^M (P \cdot S_i)_j$.
% \textcolor{red}{For example, consider a fully connected linear layer with a weight matrix $w$ of size $N \times M$ and a bias vector $\mathbf{b}$ of size $N$, where the intermediate output \( \mathbf{y} \) is given by
% $\mathbf{y} = w^{T} \mathbf{x} + \mathbf{b}$. To simulate variation in accumulation order, we use the permutation matrix $P$ to permute element-wise products $S_{i}$ before reduction where \( S_i = \{w_{i0}x_0, \dots, w_{iM}x_M\} \), computing  
% $y_i = \sum_{j=1}^M (P \cdot S_i)_j$. For the linear classifier in Sec \ref{sec:linear-classifier-synthetic}, we introduce the permutation matrix $P$ as follows: $ P \hat{\mathbf{n}} \cdot P \mathbf{x}$.} 
(3) Perform a gradient descent as specified in Eq. \ref{en:LP_optimization}, optimizing \emph{only} over permutation matrices. (4) Perform a forward pass and mark any misclassifications to generate a worst-case bound. As shown in the right panel of Fig.~\ref{fig:linear_decision_boundary_adversarial_attack}, the LP approach provides a tight bound on the \gls{EWA} attack. Next, we investigate misclassifications in GNNs, extending previous work \cite{shanmugavelu2024impactsfloatingpointnonassociativityreproducibility} which identified significant run-to-run output variability in GNNs but did not consider misclassification.%\vspace{-1em}

%%%%%%%%%%%%%%%%%%%%%%%%%%%%%%%%%%%%%%%%%%%%%%%%%%%%%%%%%%%%%%
\section{Non-Determinism in Graph Neural Networks}%\vspace{-1mm}
%%%%%%%%%%%%%%%%%%%%%%%%%%%%%%%%%%%%%%%%%%%%%%%%%%%%%%%%%%%%%%

\gls{GNNs} operate on unordered graph data. For a graph $G=(V,E)$, any permutation of $V$ and $E$ represents the same structure. \gls{GNNs} learn node and edge representations via \textit{message passing} and \textit{aggregation}, the core operations in most architectures~\cite{DBLP:journals/corr/abs-1812-08434}. Since node neighborhoods lack a fixed order, \gls{GNNs} rely on permutation-invariant aggregation like \texttt{add} and \texttt{mean} implemented in PyTorch Geometric~\cite{Torch-Documentation} with \texttt{scatter\_reduce} functions, which introduce non-determinism due to atomic operations. This, combined with the non-unique representation of graphs, makes \gls{GNNs} in PyTorch Geometric well-suited for studying \gls{APFPR} effects. On these \gls{GNNs} we investigate run-to-run variability in robustness results, identifying worst-case accuracies with the learnable permutation approach. Additionally, we perform the EWA attack (Section \ref{sec:ewa}) to induce misclassifications and evaluate the ability of the \gls{LP} approach to provide worst-case estimates.%\vspace{-1mm}

\subsection{Experimental Methodology}

We study \gls{APFPR} vulnerability in GNN architectures: GraphSAGE, GAT, and GCN~\cite{GraphSAGE,GAT,GCN}, using the CORA, CiteSeer, and PubMed datasets~\cite{CiteSeer}. These widely used benchmarks evaluate GNN performance in semi-supervised node classification. For each model-dataset pair, we analyze misclassifications due to non-deterministic functions and \gls{EWA}. We train $N_{\text{train}}=100$ models for 25 epochs, initializing them identically with fixed randomness from stochastic training and random seed settings. Training models with atomic functions have been shown to produce different weights due to \gls{APFPR}~\cite{shanmugavelu2024impactsfloatingpointnonassociativityreproducibility} and we aim to investigate the full training and inference pipeline. To assess inference variability, we perform $N_{\text{val}}=10000$ forward passes on the validation set, with and without atomics. The base class of PyTorch GNNs is non-deterministic by default~\cite{shanmugavelu2024impactsfloatingpointnonassociativityreproducibility}. To provide a deterministic control experiment, we refactor the base class with a deterministic \texttt{index\_add} operation replacing \texttt{scatter\_reduce}. However, this is \emph{not} a solution to the non-determinism problem in GNNs since significant refactoring is needed to ensure both functional parity and sufficient performance and it is unclear if this is possible. An input is marked as misclassified if any iteration produces an incorrect prediction. A large $N_{\text{val}}$ is required since misclassifications may only appear after many identical repeated runs, as shown in Section \ref{sec:linear-classifier-synthetic}. We prevent kernel switching, isolating \gls{APFPR} as the sole source of classification flips. We also predict misclassifications and worst-case accuracy bounds using the \gls{LP} method (Section~\ref{subsec:LP}) with  $1000$ optimization steps. Graph edges in PyTorch Geometric are permutation invariant, allowing us to introduce floating-point accumulation sensitivity through permutation matrices in GNN layers when passing the adjacency matrix or $\texttt{edge\_index}$ variable across layers.
\begin{table}[ht]
\centering
\caption{Average accuracy (number of correct classifications out of 500) on the CORA dataset for a 10-layer GraphSAGE model under different attacks and attack epsilon values, with standard deviation. ``ND'' and ``D'' indicate non-deterministic or deterministic PyTorch settings during inference, respectively. For $ND$, 10000 inference runs are performed. ``LP'' refers to a learnable permutation worst-case bound, determined for each input and ``EW'' refers to the external workload attack,  which succeeds at least 75\% of the time on 1000 repeated runs. We bold experiments which result in misclassifications. All experiments are performed on the H100 with default PyTorch FP32 precision.}
\begin{tabular}{|c|c|c|c|c|c|}
    \hline
    \textbf{Attack} & \textbf{Epsilon} & \textbf{Accuracy D} & \textbf{Accuracy ND} & \textbf{Accuracy LP} & \textbf{Accuracy EWA} \\ \hline
    None & $\mathbf{0}$ & $\mathbf{405} \pm \mathbf{9}$ & $\mathbf{405} \pm \mathbf{11}$ & $\mathbf{402} \pm \mathbf{8}$ & $\mathbf{403} \pm \mathbf{8}$ \\ \hline
    FGSM & $\mathbf{1e-5}$ & $\mathbf{399} \pm \mathbf{8}$ & $\mathbf{399} \pm \mathbf{8}$ & $\mathbf{397} \pm \mathbf{5}$ & $\mathbf{397} \pm \mathbf{6}$ \\ \hline
    & $1e-4$ & $397 \pm 9$ & $397 \pm 9$ & $397 \pm 9$ & $397 \pm 9$ \\ \hline
    & $1e-3$ & $394 \pm 8$ & $394 \pm 8$ & $394 \pm 8$ & $394 \pm 8$ \\ \hline
    & $1e-2$ & $369 \pm 9$ & $369 \pm 9$ & $369 \pm 9$ & $369 \pm 9$ \\ \hline
    & $\mathbf{1e-1}$ & $\mathbf{340} \pm \mathbf{9}$ & $\mathbf{340} \pm \mathbf{10}$ & $\mathbf{321} \pm \mathbf{16}$ & $\mathbf{328} \pm \mathbf{9}$ \\ 
    \hline
    PGD & $\mathbf{1e-5}$ & $\mathbf{387} \pm \mathbf{8}$ & $\mathbf{387} \pm \mathbf{8}$ & $\mathbf{385} \pm \mathbf{9}$ & $\mathbf{385} \pm \mathbf{9}$ \\ \hline
    & $1e-4$ & $365 \pm 9$ & $365 \pm 9$ & $365 \pm 9$ & $365 \pm 9$ \\ \hline
    & $1e-3$ & $348 \pm 9$ & $348 \pm 9$ & $348 \pm 9$ & $348 \pm 9$ \\ \hline
    & $\mathbf{1e-2}$ & $\mathbf{326} \pm \mathbf{9}$ & $\mathbf{326} \pm \mathbf{9}$ & $\mathbf{309} \pm \mathbf{8}$ & $\mathbf{313} \pm \mathbf{7}$ \\ \hline
    & $\mathbf{1e-1}$ & $\mathbf{301} \pm \mathbf{9}$ & $\mathbf{301} \pm \mathbf{9}$ & $\mathbf{287} \pm \mathbf{14}$ & $\mathbf{292} \pm \mathbf{15}$ \\ \hline
    Random & $\mathbf{1e-5}$ & $\mathbf{405} \pm \mathbf{8}$ & $\mathbf{405} \pm \mathbf{8}$ & $\mathbf{403} \pm \mathbf{10}$ & $\mathbf{403} \pm \mathbf{10}$ \\ \hline
    & $1e-4$ & $405 \pm 9$ & $405 \pm 9$ & $405 \pm 9$ & $405 \pm 9$ \\ \hline
    & $1e-3$ & $405 \pm 9$ & $405 \pm 9$ & $405 \pm 9$ & $405 \pm 9$ \\ \hline
    & $1e-2$ & $405 \pm 9$ & $405 \pm 9$ & $405 \pm 9$ & $405 \pm 9$ \\ \hline
    & $\mathbf{1e-1}$ & $\mathbf{402} \pm \mathbf{9}$ & $\mathbf{402} \pm \mathbf{9}$ & $\mathbf{383} \pm \mathbf{18}$ & $\mathbf{389} \pm \mathbf{20}$ \\ \hline
    Targeted & $\mathbf{1e-5}$ & $\mathbf{377} \pm \mathbf{8}$ & $\mathbf{377} \pm \mathbf{8}$ & $\mathbf{375} \pm \mathbf{9}$ & $\mathbf{375} \pm \mathbf{9}$ \\ \hline
    & $\mathbf{1e-4}$ & $\mathbf{365} \pm \mathbf{9}$ & $\mathbf{365} \pm \mathbf{9}$ & $\mathbf{359} \pm \mathbf{13}$ & $\mathbf{361} \pm \mathbf{14}$ \\ \hline
    & $\mathbf{1e-3}$ & $\mathbf{331} \pm \mathbf{9}$ & $\mathbf{331} \pm \mathbf{12}$ & $\mathbf{326} \pm \mathbf{5}$ & $\mathbf{327} \pm \mathbf{4}$ \\ \hline
    & $\mathbf{1e-2}$ & $\mathbf{316} \pm \mathbf{9}$ & $\mathbf{316} \pm \mathbf{10}$ & $\mathbf{298} \pm \mathbf{21}$ & $\mathbf{303} \pm \mathbf{21}$ \\ \hline
    & $\mathbf{1e-1}$ & $\mathbf{293} \pm \mathbf{9}$ & $\mathbf{293} \pm \mathbf{9}$ & $\mathbf{284} \pm \mathbf{15}$ & $\mathbf{288} \pm \mathbf{17}$ \\ \hline
\end{tabular}
\label{tab:accuracy_table}
%\vspace{-3mm}
\end{table}

Additionally, we performed \gls{EWA} to examine whether external workloads can induce misclassification in a real-world network. As before, we ran Bayesian optimization for 100 iterations (with 1000 experiments per iteration) to identify the optimal attack matrix size $k \in \{1000, 10000\}$ that flips classifications to the second most probable class. We then perform 1000 repeated inferences to assess the attack's success, considering it successful if misclassification occurs at least 75\% of the time. While this threshold is arbitrary, a reliable attack should consistently induce misclassification. We validate models and test  unperturbed and adversarial inputs from \gls{FGSM}, \gls{PGD}, random, and targeted attacks (Section~\ref{sec:defs}).%\vspace{-2mm}

\subsection{Results}%\vspace{-2mm}
 Results for a 10-layer GraphSAGE model on the CORA dataset are shown in Table~\ref{tab:accuracy_table}, performed on an H100. Similar behavior was observed in other datasets, models, and GPUs (details are available on our GitHub). For each adversarial attack method and epsilon value, we report test accuracy as the number of correct classifications (out of 500), with the first row showing results for $\epsilon=0$ (no attack). The Columns labeled $ND$ or $D$ indicate whether deterministic PyTorch kernels were used during inference, and the \gls{LP} column represents worst-case upper bounds determined via learned permutation optimization. The \gls{EWA} column represents the \gls{EWA} attack, which use naive matrix multiplication workloads. Errors are standard deviations over the $100$ trained models.

As expected, accuracies decrease with increasing attack strength $\epsilon$. Toggling PyTorch’s non-deterministic functions on or off has little impact on average accuracy (D and ND columns); not all PyTorch functions have a deterministic version \cite{shanmugavelu2024impactsfloatingpointnonassociativityreproducibility}. However, adversarial accuracy varies significantly at certain epsilon values, indicating that \gls{APFPR} induces additional misclassifications beyond input perturbations. Notably, large errors occur even at $\epsilon=0$, showing that non-perturbed inputs are vulnerable to \gls{APFPR}. \gls{EWA} reliably misclassifies such inputs, with targeted and PGD attacks being the most affected, leading to adversarial accuracy drops of up to 4.6\% (Targeted Attack, $\epsilon=0.01$). Random attacks are minimally impacted. The \gls{EWA} attack works $\geq$ 75\% of the time, which is at least a three-order-of-magnitude increase in run-to-run misclassification consistency. \gls{EWA} achieves convergence in less than 80 iterations during the optimization step. Tightening the constraint $\geq$ 85\% makes  \gls{EWA} ineffective and relaxing it $\leq$ 20\% makes it always effective, producing accuracy values closer to the \gls{LP} approach. The \gls{LP} method provides strong worst-case bounds and should be integrated into existing robustness verification tools as a workload as simple and common as a matrix multiplication in \gls{EWA} is sufficient to induce misclassifications in GNNs and simpler linear classifiers as in Sec.~\ref{sec:ewa}. As before, an erratic trend was found between optimal matrix size and inputs. We leave an in-depth analysis to future work, requiring the development of novel GPU scheduler probing tools. %\vspace{-3mm}

\section{Impact of GPU State on Order of Reductions} \label{sec:gpu-state}%\vspace{-2.5mm}

The previous section demonstrates that GPU states can significantly impact \gls{ML} workloads and should be considered as seriously as more conventional attacks. However, it only provides an indirect measure of the mechanisms behind misclassifications. In this section, we directly measure how the GPU state affects the ordering of asynchronous parallel operations, such as those involving CUDA’s \atomicadd. While exact scheduler behavior is not always understood \cite{otterness2021exploring, Olmedo2020-nvidia-scheduler}, the scheduler plays a critical role in determining the order of these operations. To our knowledge, no prior studies have specifically explored how the GPU state influences the ordering of asynchronous operations.%\vspace{-3mm}

\subsection{Methodology}%\vspace{-1mm}

We use the parallel sum algorithm to show how the asynchronous operation order, measured by \gls{BIEO}, is influenced by external loads. The reduction $\sum_i^n x_i$, where $x_i\neq x_j\ \forall i,j,\ x_i>0$ \gls{FP64} numbers, runs on a GPU using \atomicadd, which has an undefined execution order. To recover the execution order, we track the accumulator updates per block and sort them post-execution.
We sum 100 lists of 1M uniform \gls{FP64} numbers, with and without an additional \gls{DGEMM} workload. Each test was run 10 times to account for system variations. Sorting yields two datasets—\gls{RO} and \gls{RDGEMM}. The Kendall’s $\tau$ correlation~\cite{kendall1938new} measures permutation similarity: \kRO\ for \gls{RO} and \kRORDGEMM\ comparing \gls{RO} to \gls{RDGEMM}. 
%Further work probing the GPU scheduler is necessary to investigate reductions of more general arrays.\vspace{-1mm}
This method shows the GPU states' effects on the execution order, aiding with verification in non-deterministic settings. We tested various GPUs (Section \ref{sec:gpu-details}), power settings, and partitions, presenting results for GH200 and V100.%\vspace{-1em}

%\subsection{Impact of GPU states on atomic operations order}

\subsection{Results}

\subsubsection{Impact of External Workloads} 
\begin{figure}[h!]
\centering 
\includegraphics[width=0.45\textwidth]{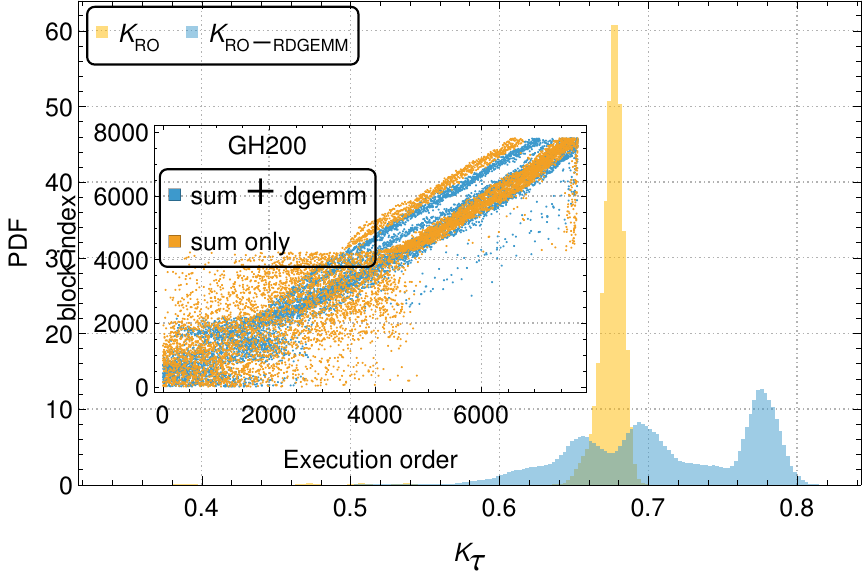}
\includegraphics[width=0.45\textwidth]{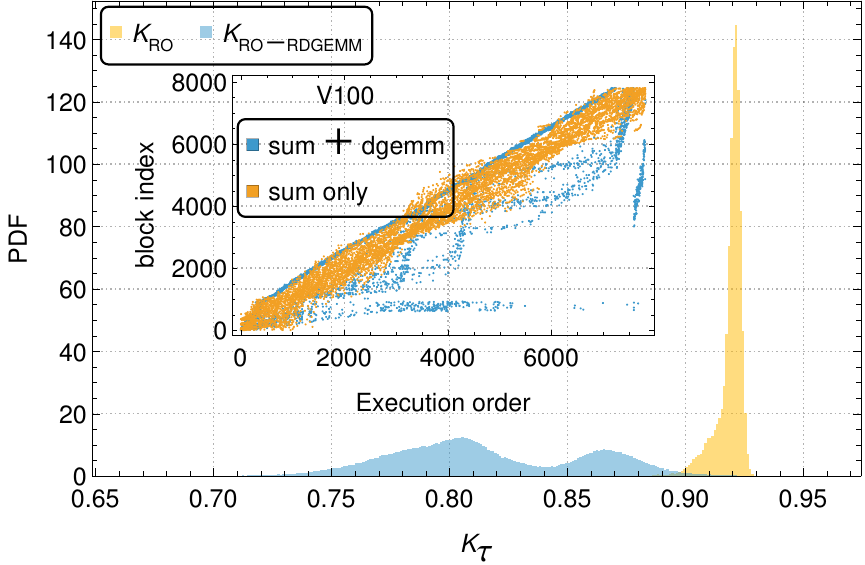}
\caption{\textbf{Left panel:} \gls{PDF} \kRO\ and \kRORDGEMM\ on GH200. \textbf{Right panel:} \kRORDGEMM\ \gls{PDF} \kRO\ and \kRORDGEMM\ on V00. The inset in the figures shows \gls{BIEO} for the \gls{RO} and \gls{RDGEMM} workloads with the lowest \kRORDGEMM\ correlation.} 
\label{fig:GPU:KendallTau} 
%\vspace{-1em}
\end{figure}

We calculated \kRO\ and \kRORDGEMM\ for the GH200 GPU architecture. As shown in Fig.~\ref{fig:GPU:KendallTau}, GH200 exhibits distinct block scheduling and atomic instruction behaviors. The \kRO\ distribution ranges from $0.32$ to $0.70$, peaking at $0.67$, while \kRORDGEMM\ ranges from $0.45$ to $0.83$ with a mean of $0.71$. The multimodal distributions for \gls{DGEMM} workloads differ significantly from the unimodal distributions for unperturbed reductions, reflecting the sensitivity to external workloads. 

As shown in Fig.~\ref{fig:GPU:KendallTau}, the V100 GPU behaves differently, as \kRO\ and \kRORDGEMM\ distributions are narrower than on GH200. The \kRORDGEMM\ distribution remains multimodal but has a more complex structure than on the GH200 GPU. The inset highlights the \gls{BIEO} snapshot for the pair with the lowest Kendall $\tau$ correlation, showing non-sequential block index ordering at first, which later converges into two parallel distributions. These results show that external workloads significantly influence the execution order of \atomicadd\ on GH200, expanding the range of possible ordering permutations.%\vspace{-3mm}

\subsubsection{Impact of MiG Configuration} 
\begin{figure}[htbp]
\centering
\includegraphics[width=0.45\textwidth]{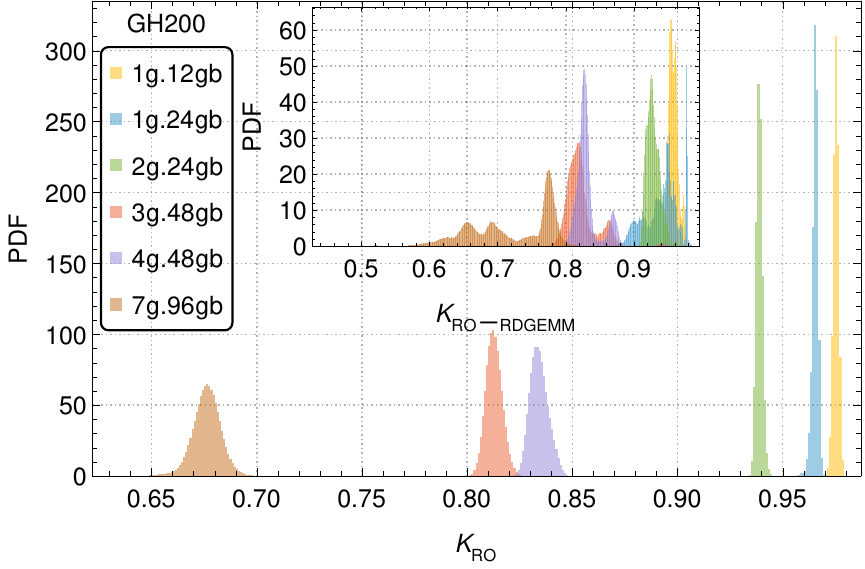}
\includegraphics[width=0.45\textwidth]{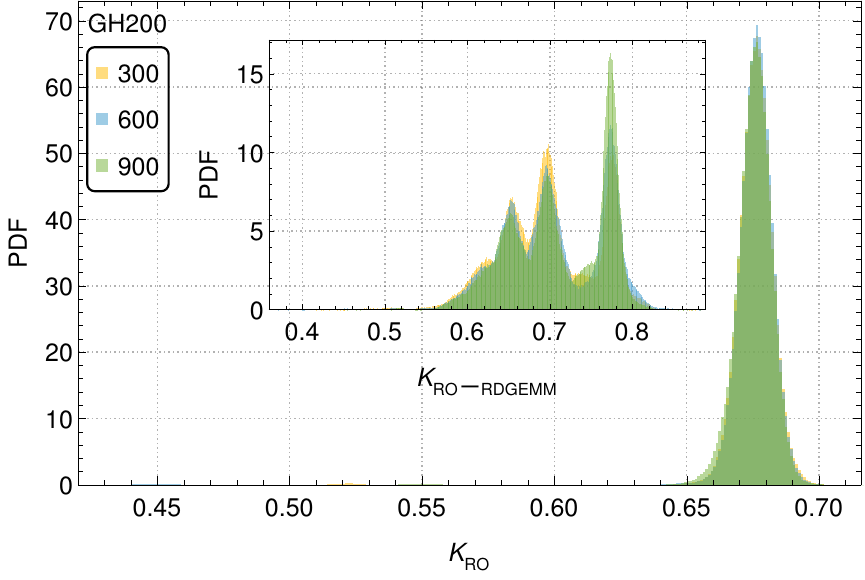}
\caption{\textbf{Left panel:} \kRORDGEMM\,  \gls{PDF} for six different \gls{MiG} configurations on GH200 showing the effect of resource restrictions on the Kendall $\tau$ correlations. \textbf{Right panel:} impact of power capping on the Kendall $\tau$ correlations.}
\label{fig:power:gh200:MIG}
%\vspace{-3mm}
\end{figure}

The GH200 supports up to seven \gls{MiG} partitions, allowing resource sharing. The left panel of Fig.~\ref{fig:power:gh200:MIG} shows \kRORDGEMM\ across \gls{MiG} configurations. Higher values occur with fewer SM units and decrease as resources increase, while wider distributions in larger GPUs suggest greater permutation variability. Since MiG partitions share the same hardware, each slice's behavior depends on the runtime configuration. This is especially relevant in virtualized cloud environments.%\vspace{-4mm}

%\vspace{-3mm}

\subsubsection{Impact of Power}

Research shows that power capping improves energy efficiency and reduces hardware failure rates in HPC and DL workloads~\cite{zhao2023sustainable}. Modern GPUs, such as the NVIDIA GH200, include power capping to regulate power draw~\cite{zhao2023sustainable}. While power capping reduces GPU clock speeds, increasing scheduling latency and slowing thread execution, it can also lead to resource contention and impact the atomic operation ordering in compute-intensive tasks. As shown in the right panel of Fig.\ref{fig:power:gh200:MIG} we observed no significant differences in the \gls{PDF} of \kRO\ and \kRORDGEMM, suggesting that power capping does not notably affect instruction ordering.%\vspace{-3mm}

\section{Discussion and Conclusions} %\vspace{-2mm}
We show that \gls{APFPR} has significant impacts on classification accuracy and robustness. We developed a novel black-box Bayesian optimization attack (\gls{EWA}) to determine the properties of additional workloads that reliably result in misclassification (up to 4.6\% accuracy decrease, at least 75\% of the time). While our current work focused on matrix multiplications as the external background workload, future research should explore more complex workloads in both isolated and cloud environments. Additionally, GPU scheduler probing tools must be developed to investigate how \gls{EWA} and associated workloads impact misclassification. We introduced the \gls{LP} approach to efficiently identify permutations that maximize prediction errors, offering significant advantages over brute-force search. Our results demonstrate that both run-to-run variability and \gls{EWA} can be bound by the LP worst-case estimates. Direction for future work involves optimizing LP further, integrating it into existing robustness verification tools and performing more exhaustive testing over different ML architectures and datasets.

Our examination of GPU system states, including varying workloads, partitions, and power settings, showed a significant influence on the ordering of parallel operations across three different GPU models (from NVIDIA and AMD). These findings highlight that testing with a single GPU type is insufficient to fully account for non-determinism in model performance. This further emphasizes the value of our LP approach over repeated inferences, which would otherwise require extensive testing across multiple GPUs and GPU states.
While frameworks like PyTorch can support deterministic operations, non-deterministic kernels and atomic operations are deeply integrated, making full determinism costly to implement. Workarounds such as integer quantization may help but often reduce accuracy, particularly in deep architectures. Purpose-built deterministic hardware such as the Groq LPU offers a valuable alternative benchmark for reliable inference. %\vspace{-1em}

%While deterministic kernels are technically possible in frameworks like PyTorch, non-deterministic kernels and atomic operations are deeply integrated into ML frameworks, requiring substantial refactoring to achieve determinism. Workarounds, including integer quantization, may help, but potentially at the cost of accuracy degradation, particularly for deep architectures. Additionally, deterministic architectures like the Groq LPU may provide a valuable baseline. \vspace{-1em}
% reference point. %for evaluating non-determinism susceptibility in model performance.

\section{Hardware and Systems Used in Experiments} \label{sec:gpu-details} %\vspace{-3mm}

Tests for the V100 are run on the Summit supercomputer at the Oak Ridge Leadership Computing Facility (OLCF), running Redhat OS 8. Summit is an IBM system; each IBM Power System AC922 node has two Power9 CPUs with 512 GB of memory and 6 V100 NVIDIA GPU with 16GB of HBM2 memory. 

Tests on Mi250X AMD GPU are obtained on the Frontier supercomputer at OLCF, running SLE 15 (enterprise). Frontier is an HPE Cray EX supercomputer; each Frontier compute node has a 64-core AMD ``Optimized 3rd Gen EPYC” CPU with 512 GB of DDR4 memory and 4 AMD MI250X GPUs, each with 2 Graphics Compute Dies (GCDs) for a total of 8 GCDs per node. 

Tests on GH200 GPUs are run on two separate compute nodes, one running SLE 15 (enterprise) and the other Red Hat Enterprise Linux 9.4 (Plow) with 2 NVIDIA GH200 GPUs and 72-core ARM Neoverse-V2 CPUs. H100 tests run on Ubuntu 22.04.06 with two 40GB H100 GPUs and an AMD EPYC 7302 CPU. We use PyTorch 2.4, PyTorch Geometric 2.6 and CUDA 12.0.%\vspace{-3mm}

\section*{Acknowledgments}%\vspace{-2mm}
 This work was supported in part by the ORNL AI LDRD Initiative, the Swiss Platform For Advanced Scientific Computing (PASC), and the Accelerated Data Analytics and Computing Institute (ADAC). It used resources of the OLCF, a DOE Office of Science User Facility [DE-AC05-00OR22725], and the Swiss National Supercomputing Centre. The authors thank Hayashi Akihiro and Pim Witlox for insightful discussions.
 
%\vspace{-3mm}
\section*{Disclosure of Interests}%\vspace{-2mm}
The authors have no competing interests to declare that are relevant to the content of the article. 
%\vspace{-3mm}
\bibliographystyle{splncs04}% \vspace{-2mm}
\bibliography{references}
\end{document}